\documentclass[11pt,a4paper]{article}
\usepackage[hyperref]{AACL-IJCNLP2020}
\usepackage{times}
\usepackage{latexsym}
\usepackage{url}
\usepackage{flushend}
\usepackage{graphicx}
\usepackage{amsmath}
\usepackage{subcaption}
\usepackage{amsfonts}
\usepackage{tabulary,booktabs}
\usepackage[export]{adjustbox}

\usepackage{microtype}

\aclfinalcopy % Uncomment this line for the final submission
\def\aclpaperid{5} %  Enter the aacl-ijcnlp Paper ID here

\title{ISA: An Intelligent Shopping Assistant}

\author{Tuan Manh Lai \footnotemark\, \textsuperscript{1,2}, Trung Bui \textsuperscript{2}, Nedim Lipka \textsuperscript{2}  \\
\textsuperscript{1} University of Illinois at Urbana-Champaign\\
\textsuperscript{2} Adobe Research
}

\begin{document}
\maketitle
\addtocounter{footnote}{1}
\footnotetext{\,The work was conducted while the first author interned at Adobe Research.}
\begin{abstract}
Despite the growth of e-commerce, brick-and-mortar stores are still the preferred destinations for many people. In this paper, we present ISA, a mobile-based intelligent shopping assistant that is designed to improve shopping experience in physical stores. ISA assists users by leveraging advanced techniques in computer vision, speech processing, and natural language processing. An in-store user only needs to take a picture or scan the barcode of the product of interest, and then the user can talk to the assistant about the product. The assistant can also guide the user through the purchase process or recommend other similar products to the user. We take a data-driven approach in building the engines of ISA's natural language processing component, and the engines achieve good performance.
\end{abstract}

\section{Introduction}
Shopping in physical stores is a popular option for many people. Each week, a lot of people enter supermarkets in which they are immersed with many different product choices. In many shopping centers, customer service representatives (CSRs) are employed to answer questions from customers about products. However, a customer may experience long waiting time for assistance if all CSRs are busy interacting with other customers. Therefore, automated solutions can increase customer satisfaction and retention.

In this paper, we introduce a mobile-based intelligent shopping assistant, ISA, which is based on advanced techniques in computer vision, speech processing, and natural language processing. A user just needs to take a picture or scan the barcode of the product of interest. After that, the user can ask ISA a variety of questions such as product features, specifications and return policies. The assistant can also guide the user through the purchase process or recommend other similar products. This work can be used as the first step in fully automating customer service in shopping centers. With ISA, no CSRs will be needed as customers can simply turn to their phones for assistance. We have developed a fully functional prototype of ISA.

The rest of the paper is organized as follows. Section 2 introduces some related work. Section 3 gives an overview of the design and implementation of the system. Finally, Section 4 concludes the paper and suggests future directions.

\section{Related Work}
\begin{figure}
  \centering
  \includegraphics[width=\linewidth]{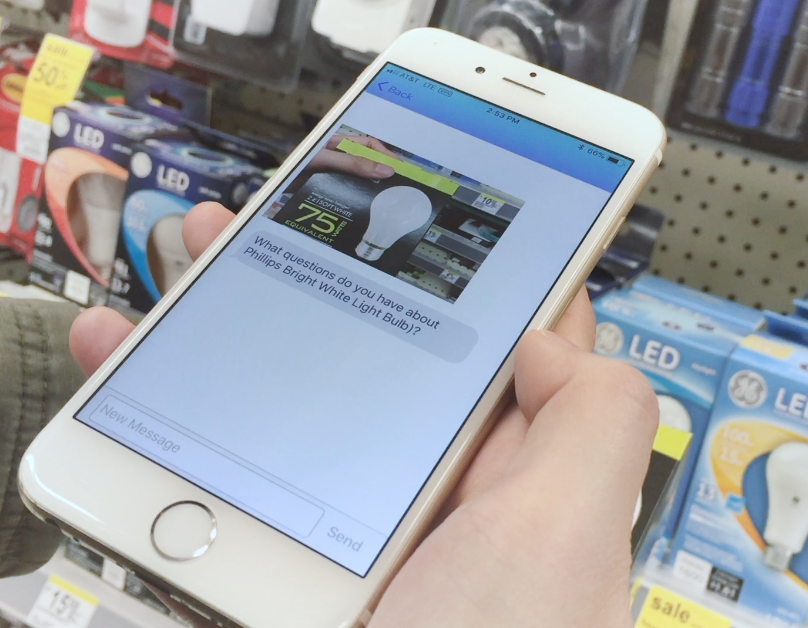}
  \caption{ISA assists users at physical stores}
  \label{fig:at_physical_store}
\end{figure}

\begin{figure*}[!htb]
  \centering
  \includegraphics[width=\linewidth]{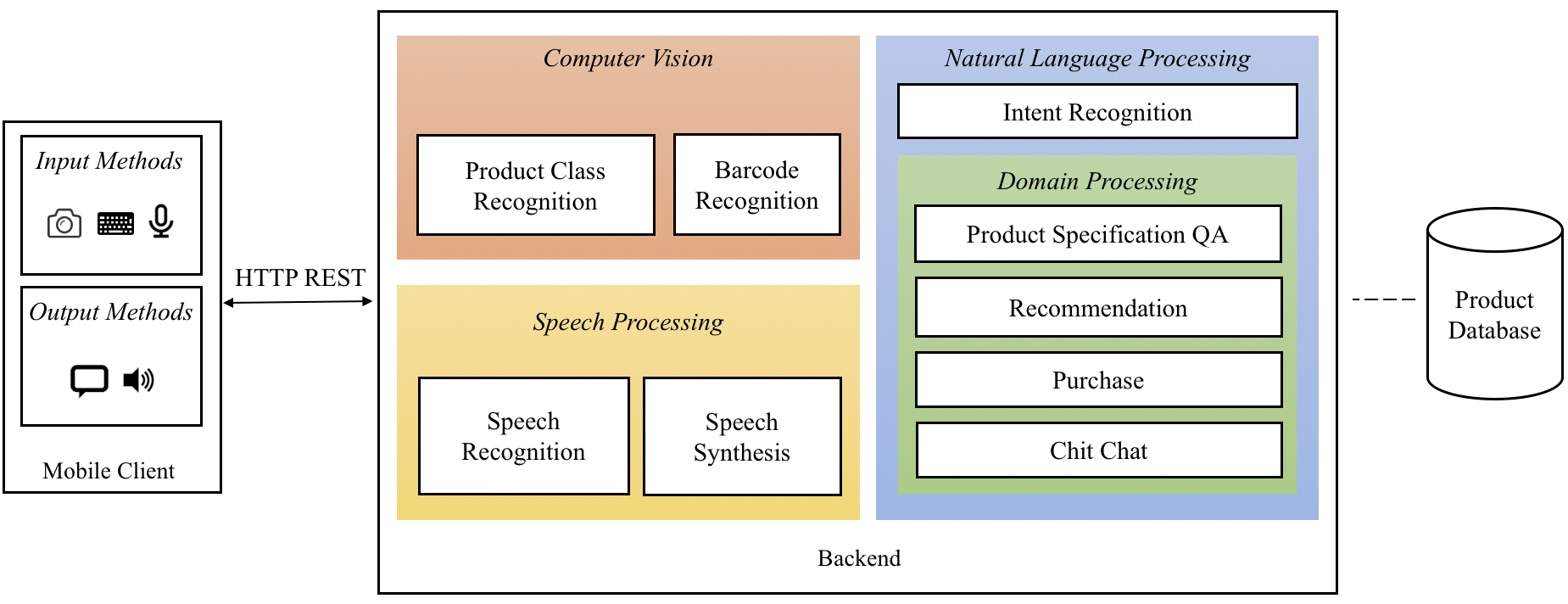}
  \caption{The system overview of ISA}
  \label{fig:system_overview}
\end{figure*}

The most closely related branches of work to ours are probably customer service chatbots for e-commerce websites. For example, SuperAgent \cite{superagent-customer-service-chatbot-e-commerce-websites} is a powerful chatbot that leverages large-scale and publicly available e-commerce data. The researchers demonstrate SuperAgent as an add-on extension to mainstream web browsers. When a user visits a product page, SuperAgent crawls the information of the product from multiple data sources within the page. After that, the user can ask SuperAgent about the product. Unlike SuperAgent, ISA is designed to assist users at physical stores (Figure \ref{fig:at_physical_store}). In addition to natural language processing techniques, ISA also needs to use techniques in computer vision and speech processing when interacting with the users.

\section{System Description}
\subsection{Overview}
When an in-store user wants to get more information about a specific product, the user just needs to take a picture or scan the barcode of the product. The system then retrieves the information of the product of interest from a database by using computer vision techniques. After that, the user can ask natural language questions about the product specifications to the system. The user can either type in the questions or directly speak out the questions using voice. ISA is integrated with both speech recognition and speech synthesis abilities, which allows users to ask questions without typing.

Figure \ref{fig:system_overview} shows the system overview of ISA. As the figure shows, a mobile client communicates with the backend through a well-defined HTTP REST API. This creates a separation between the client and the server, which allows ISA to be scaled without much difficulty. The backend consists of three main components: 1) speech processing, 2) computer vision, 3) natural language processing. Users can chat with ISA in speech. The speech recognition and speech synthesis are implemented by calling third-party services. The computer vision component is responsible for recognizing the products that the user is facing. Given an image of a product of interest, a fine-grained visual object classification model will be used to identify the product and retrieve its information. This task is challenging because many products are visually very similar (e.g., washers and dryers usually have similar shape). Therefore, we enhance the component with highly accurate standard algorithms for barcode recognition. In case it is difficult for the object classification model to identify the product of interest accurately, the user can simply scan the barcode of the product. Finally, the natural language processing component is responsible for generating a response from  a text query or question. We will next detail each part of the natural language processing component in the following sections.

\begin{table}
\centering
\small
\begin{tabular}{|c|c|}
\hline \bf Intent Types & \bf Example Query \\ \hline
Product Specification QA & How heavy is this chair? \\ \hline
Recommendation & Show me some other items \\ \hline
Purchase & I want to buy this. \\ \hline
Chit Chat & How are you doing? \\
\hline
\end{tabular}
\caption{\label{intent_types} Intent Types }
\end{table}

\subsection{Intent Recognition}
When ISA receives a query from a user, the intent recognition engine is used to determine the intent of the query. Based on the recognized intent, the appropriate domain-specific engine will be triggered. We define four different types of intent as shown in Table \ref{intent_types}. Intent detection can be naturally treated as a classification problem. In this work we build a random forest model \cite{Breiman:2001:RF:570181.570182} for the problem and it achieves good performance. Other popular classifiers like support vector machines \cite{svm_complex_call_classification} and deep neural network methods \cite{deep_belief_nets_for_call_routing} can also be applied in this case. 

We create a dataset of 500 different queries and use it to build a random forest (RF) for intent classification. Approximately 2/3 of the cases are used as training set, whereas the rest (1/3) are used as test set, in order to estimate the model's performance. We create a bag-of-words feature vector for each query and use it as input for the RF. The number of trees in the forest is set to be 80. For each node split during the growing of a tree, the number of features used to determine the best split is set to be  \(\,\sqrt[]{k}\) where \(k\) is the total number of features of the dataset. The accuracy of the trained RF model evaluated on the test set is 98.20\%.

\begin{figure}
  \centering
  \includegraphics[width=\linewidth,height=8.5cm,keepaspectratio,frame]{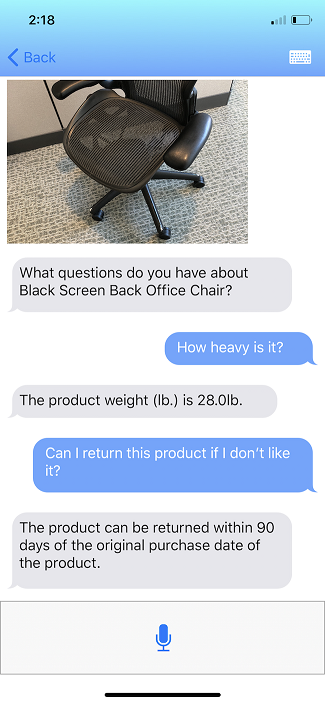}
  \caption{Answering questions regarding product specifications}
  \label{fig:product_qa_output}
\end{figure}

\begin{figure}[!htb]
  \centering
  \includegraphics[width=\linewidth,height=8.5cm,keepaspectratio,frame]{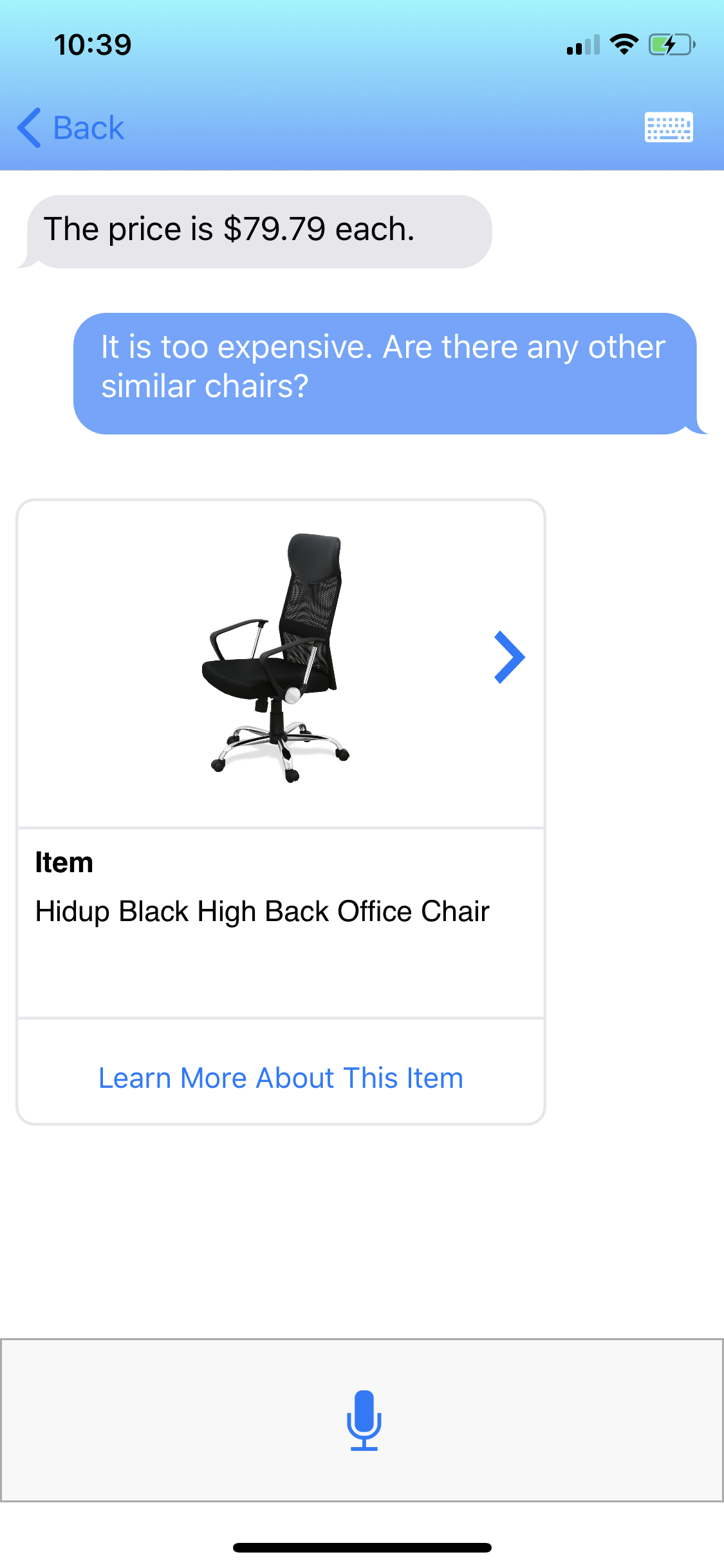}
  \caption{ISA recommends similar products to the user}
  \label{fig:recommendation_process}
\end{figure}

\subsection{Product Specification QA}
The product specification QA engine is used to answer questions regarding the specifications of a product. For every product, there is a list of specifications in the form of (specification\_name, specification\_value). We formalize the task of the engine as follows: Given a question \(Q\) about a product \(P\) and the list of specifications \((s_{1}, s_{2}, ..., s_{M})\) of \(P\), the goal is to identify the specification that is most relevant to the question \(Q\). \(M\) is the number of specifications of the product, and \(s_{i}\) is the sequence of words in the name of the \(i^{th}\) specification. In this formulation, the task is similar to the answer selection problem. `Answers' shall be individual product specifications. 

Previous methods for answer selection typically relies on feature engineering, linguistic tools, or external resources \cite{Wang:2010:PTM:1873781.1873912,Heilman:2010:TEM:1857999.1858143,question-answering-using-enhanced-lexical-semantic-models,Yao13answerextraction}. Recently, with the renaissance of neural network models, many deep learning based methods have been proposed to tackle the answer selection problem~\cite{rao2016noise,wang2017bilateral,bian17compare,shen2017inter,TranEtal2018Context,LaiEtal2018Simple,Tay2018MultiCastAN,lai2018supervised,LaiEtal2018Review,rao-etal-2019-bridging,lai-etal-2019-gated,garg2019tanda,Kamath2019PredictingAI,laskar-etal-2020-contextualized}. These deep learning based methods typically outperform traditional techniques without relying on any feature engineering or expensive external resources. For example, the IWAN model proposed in~\cite{shen2017inter} achieves competitive performance on public datasets such as TrecQA~\cite{Wang2007WhatIT} and WikiQA~\cite{wikiqa-a-challenge-dataset-for-open-domain-question-answering}.

Using Amazon Mechanical Turk, a popular crowdsourcing platform, we create a dataset of 6,922 questions that are related to 369 specifications and 148 products listed in the Home Depot website. We implement the IWAN model and train the model on the collected dataset. The top-1 accuracy, top-2 accuracy, and top-3 accuracy of the model evaluated on a held-out test set are 85.60\%, 95.80\%, and 97.60\%, respectively.

In production, given a question about a product, the trained model is used to rank every specification of the product based on how relevant the specification is. We select the top-ranked specification and use it to generate the response sentence using predefined templates \cite{superagent-customer-service-chatbot-e-commerce-websites}. An example of the product specification QA engine's outputs is shown in Figure \ref{fig:product_qa_output}. The first question from the user is matched to the product weight specification, whereas the second question is matched to the return policy specification.

\subsection{Recommendation}
The recommendation engine is responsible for giving new suggestions and recommendations to users. When a user wants to look for similar products (e.g., by saying ``Are there any other similar products?"), the engine will search the database for related products and then send the information of them to the app for displaying to the user (Figure \ref{fig:recommendation_process}).

\subsection{Purchase}
\begin{figure}
  \centering
  \includegraphics[width=\linewidth,height=8.5cm,keepaspectratio]{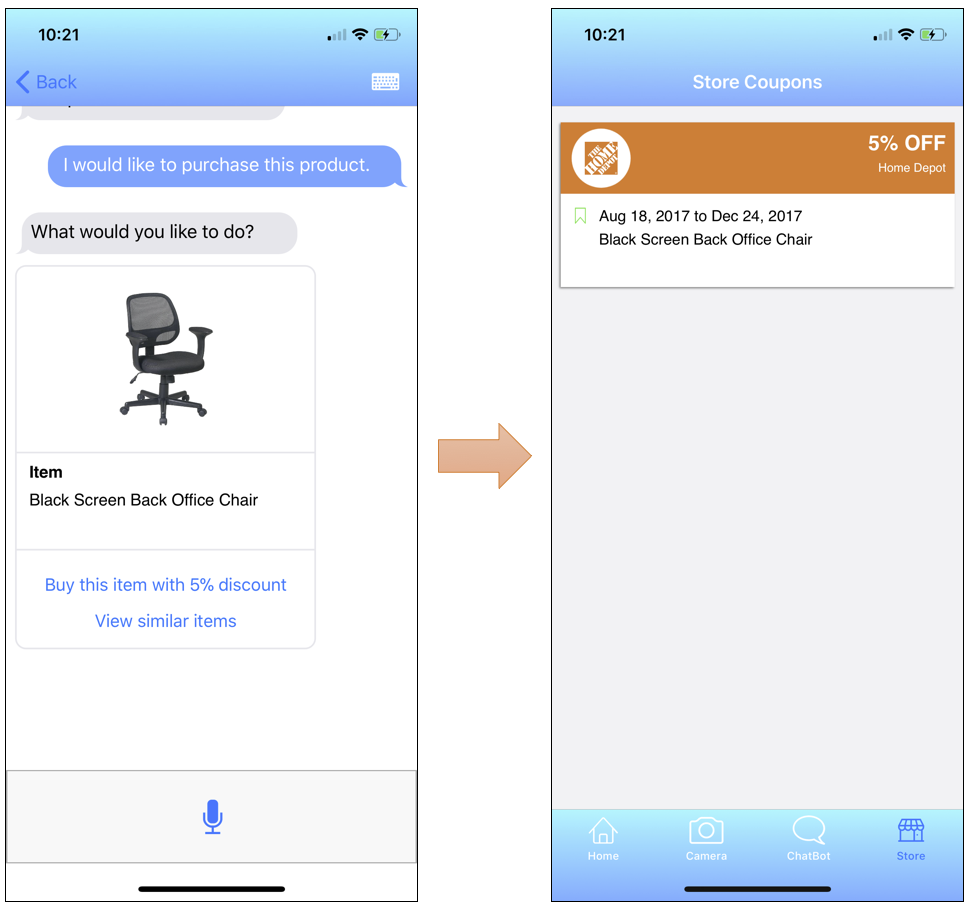}
  \caption{The user purchased an office chair with 5\% discount}
  \label{fig:purchase_process}
\end{figure}

The purchase engine is responsible for guiding the user through the purchase process. When a user wants to buy a specific product (e.g., by saying ``I would like to purchase this product."), the engine will first query the database for information such as the product listing price, available discounts, and user payment information. After that, the engine will craft a special response message and send it to the client app in the user's mobile device. The response message will instruct the app how to assist the user through the purchase process or provide personalized discounts if applicable (Figure \ref{fig:purchase_process}).

\subsection{Chit Chat}
The chit chat engine is used to reply to greeting queries such as ``How are you doing?'' or queries that are off the subject such as ``Is the sky blue?''. Our approach to building the engine is based on the sequence-to-sequence (seq2seq) framework \cite{Sutskever2014SequenceTS}. The model consists of two recurrent neural networks: an encoder and a decoder. The encoder converts the input query into a fixed size feature vector. Based on that feature vector, the decoder generates the output response, one word at a time. The model is integrated with the global attention mechanism \cite{LuongPM15} so that the decoder can attend to specific parts of the input query when decoding instead of relying only on the fixed size feature vector. We collect about 3M query-response pairs from Reddit and use them to train the seq2seq model. Examples of the engine’s outputs are
shown below:
\newline
\newline
\textbf{Q}: How are you doing?
\newline
\textbf{A}: I'm doing well.
\newline
\textbf{Q}: Is the sky blue?
\newline
\textbf{A}: Yes.

\section{Conclusion and Future Work}
In this paper, we present ISA, a powerful intelligent shopping assistant. ISA is designed to achieve the goal of improving shopping experience in  physical stores by leveraging advanced techniques in computer vision, speech processing, and natural language processing. A user only needs to take a picture or scan the barcode of the product of interest, and then the user can ask ISA a variety of questions about the product. The system can also guide the user through the purchase decision or recommend other similar products to the user.

There are many fronts on which we will be exploring in the future. Currently the product specification QA engine answers only questions regarding the specifications of a product. We will implement engines for addressing other kinds of questions. We will also extend ISA to better support other languages and informal text \cite{phobert,bertweet,2020camembert}. In addition, we will conduct a user study to evaluate our system in the future. Finally, we wish to extend this work to other domains such as building an assistant for handling image editing requests \cite{brixey2018system}.

\section{Acknowledgments}
The authors wish to thank Dr. Hung Bui (VinAI Research) and Dr. Sheng Li (University of Georgia) for their guidance and feedback on this project.

\bibliography{anthology,aacl-ijcnlp2020}
\bibliographystyle{acl_natbib}

\end{document}